\documentclass[10pt,twocolumn,letterpaper]{article}

\usepackage{iccv}
\usepackage{times}
\usepackage{epsfig}
\usepackage{graphicx}
\usepackage{amsmath}
\usepackage{amssymb}

\usepackage{multirow} 
\usepackage{array}
\usepackage{subfigure}
\usepackage{cite}
\usepackage{stfloats}
\usepackage{bbding} 
\usepackage{amssymb}
\usepackage{colortbl} 
\usepackage{bm} 

\usepackage{color}
\definecolor{MyCyan}{RGB}{224,255,255}

\usepackage[breaklinks=true,bookmarks=false,colorlinks=true,pagebackref=true,linkcolor=red,citecolor=green,]{hyperref}

\iccvfinalcopy % *** Uncomment this line for the final submission

\ificcvfinal\fi

\begin{document}

%%%%%%%%% TITLE
\title{SIENet: Spatial Information Enhancement Network for 3D Object Detection from Point Cloud}

\author{
	Ziyu Li$^{1}$*\quad
	Yuncong Yao$^{1}$\quad
	Zhibin Quan$^{1}$\quad
	Wankou Yang$^{1}$\quad
	Jin Xie$^{2}$\\
$^{1}$ Southeast University, School of Automation\\
$^{2}$ Nanjing University of Science and Technology, School of Computer Science and Engineering\\
}

\maketitle
{\let\thefootnote\relax\footnote{*Codes will be publicly available at \url{https://github.com/Liz66666/SIENet}. E-mail: liziyu@seu.edu.cn}}
\ificcvfinal\thispagestyle{empty}\fi

%%%%%%%%% ABSTRACT
\begin{abstract}
   LiDAR-based 3D object detection pushes forward an immense influence on autonomous vehicles. Due to the limitation of the intrinsic properties of LiDAR, fewer points are collected at the objects farther away from the sensor. This imbalanced density of point clouds degrades the detection accuracy but is generally neglected by previous works. To address the challenge, we propose a novel two-stage 3D object detection framework, named SIENet. Specifically, we design the Spatial Information Enhancement (SIE) module to predict the spatial shapes of the foreground points within proposals, and extract the structure information to learn the representative features for further box refinement. The predicted spatial shapes are complete and dense point sets, thus the extracted structure information contains more semantic representation. Besides, we design the Hybrid-Paradigm Region Proposal Network (HP-RPN) which includes multiple branches to learn discriminate features and generate accurate proposals for the SIE module. Extensive experiments on the KITTI 3D object detection benchmark show that our elaborately designed SIENet outperforms the state-of-the-art methods by a large margin.
\end{abstract}

%%%%%%%%% BODY TEXT
\section{Introduction}
\label{sec_introduction}

\begin{figure}[t]
	\centering	
	\includegraphics[width=\linewidth]{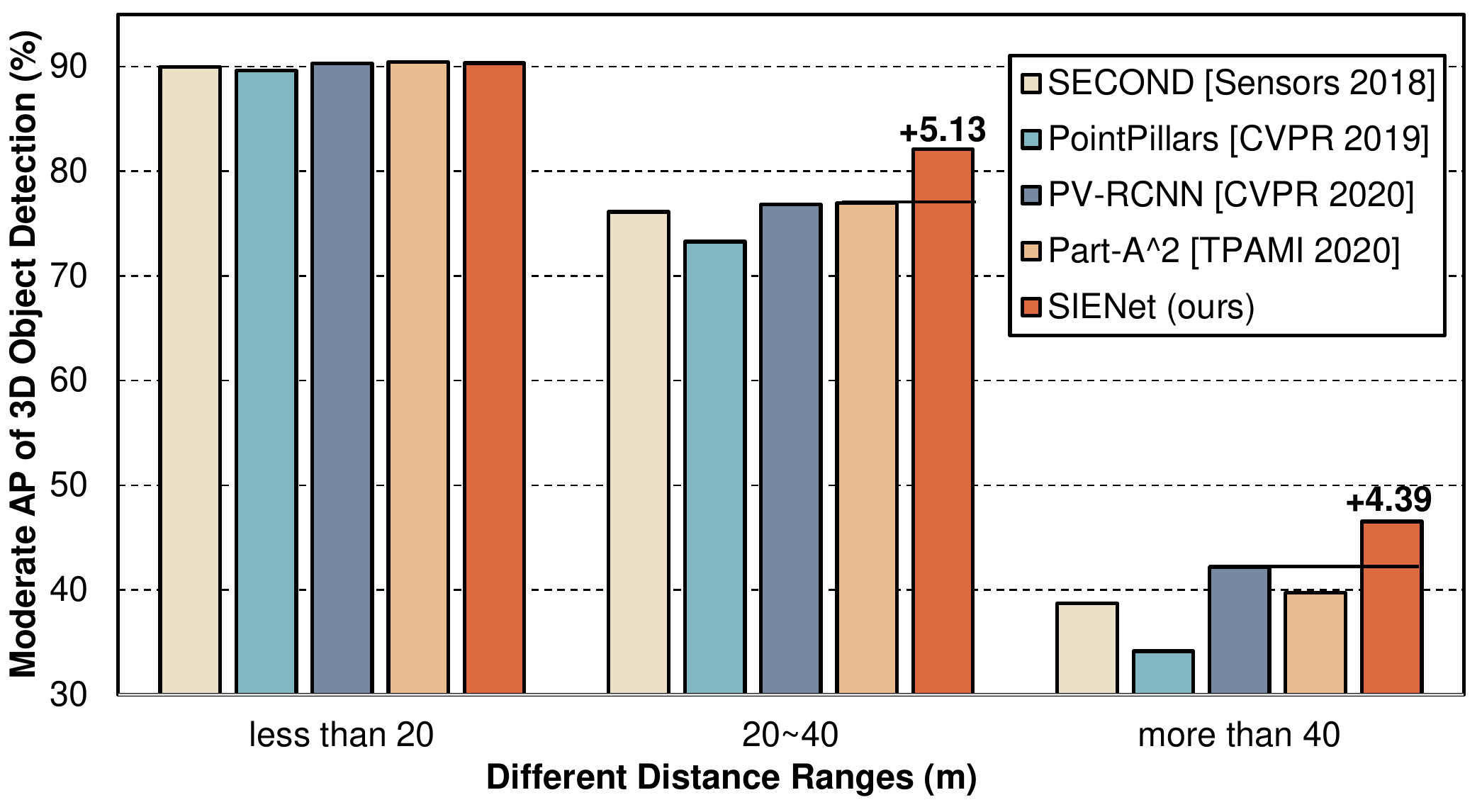}
	\caption{Detection performance at different distance ranges. Our SIENet outperforms previous state-of-the-art methods by a large margin when the distance is greater than $ 20m $.}
	\label{fig_range_results_3d}
\end{figure}

The past few decades have witnessed remarkable progress of deep learning on 2D vision tasks, such as detection \cite{faster-rcnn, ssd, yolo}, segmentation \cite{mask-rcnn, fcn, deeplab} and pose estimation \cite{hourglass, cpn, hrnet}. In contrast to 2D scenes, 3D object detection has drawn much more attention recently and brought extensive applications, including autonomous vehicles, augmented reality and indoor navigation \cite{kitti}. Compared with images, point clouds contain more geometric and semantic information, thus they become the input data of mainstream 3D object detection methods \cite{pointrcnn, std, votenet, voxelnet, pointpillars, second, part, pvrcnn}. Although point clouds could provide adequate information for neural networks to localize objects in 3D scenes, there are still many challenges in feature extraction and representation, \textit{i.e.}, unordered, sparse and locally interactive~\cite{pointnet}. Thus, how to extract point cloud features has become the principal issue for 3D object detectors.

According to the feature representation, most existing 3D detectors could be divided into two categories, \textit{i.e.}, multi-modality fusion \cite{mv3d, avod, epnet, 3dcvf,frustum} and LiDAR-based methods \cite{voxelnet, second, part, pointrcnn, votenet, std, 3dssd, pvrcnn}. Concretely, multi-modality fusion methods, such as MV3D \cite{mv3d} and AVOD \cite{avod}, first project point clouds into different views and then fuse the representations under distinct modalities. Nevertheless, this projection will lose spatial structure information and seriously decrease the inference speed, hence the majority of current works have adopted LiDAR-based frameworks, which include voxel-based \cite{voxelnet, second, part} and point-based methods \cite{pointrcnn, votenet, std, 3dssd}. For the voxel-based methods, VoxelNet \cite{voxelnet} proposes to equally divide point cloud into 3D space voxels, and stack 3D convolution layers to learn spatial features. Based on this framework, SECOND \cite{second} designs the sparse 3D convolution layer to further accelerate training and inference. Meanwhile, with the appearance of PointNet \cite{pointnet} and PointNet++ \cite{pointnet++}, point-based methods gradually become popular. PointRCNN \cite{pointrcnn} is an effective point-based detector, which utilizes PointNet++ to learn discriminant features for point-wise segmentation and proposal generation, while the back-end network refines the proposals and predicts the bounding boxes.

\begin{figure}[t]
	\centering	
	\includegraphics[width=\linewidth]{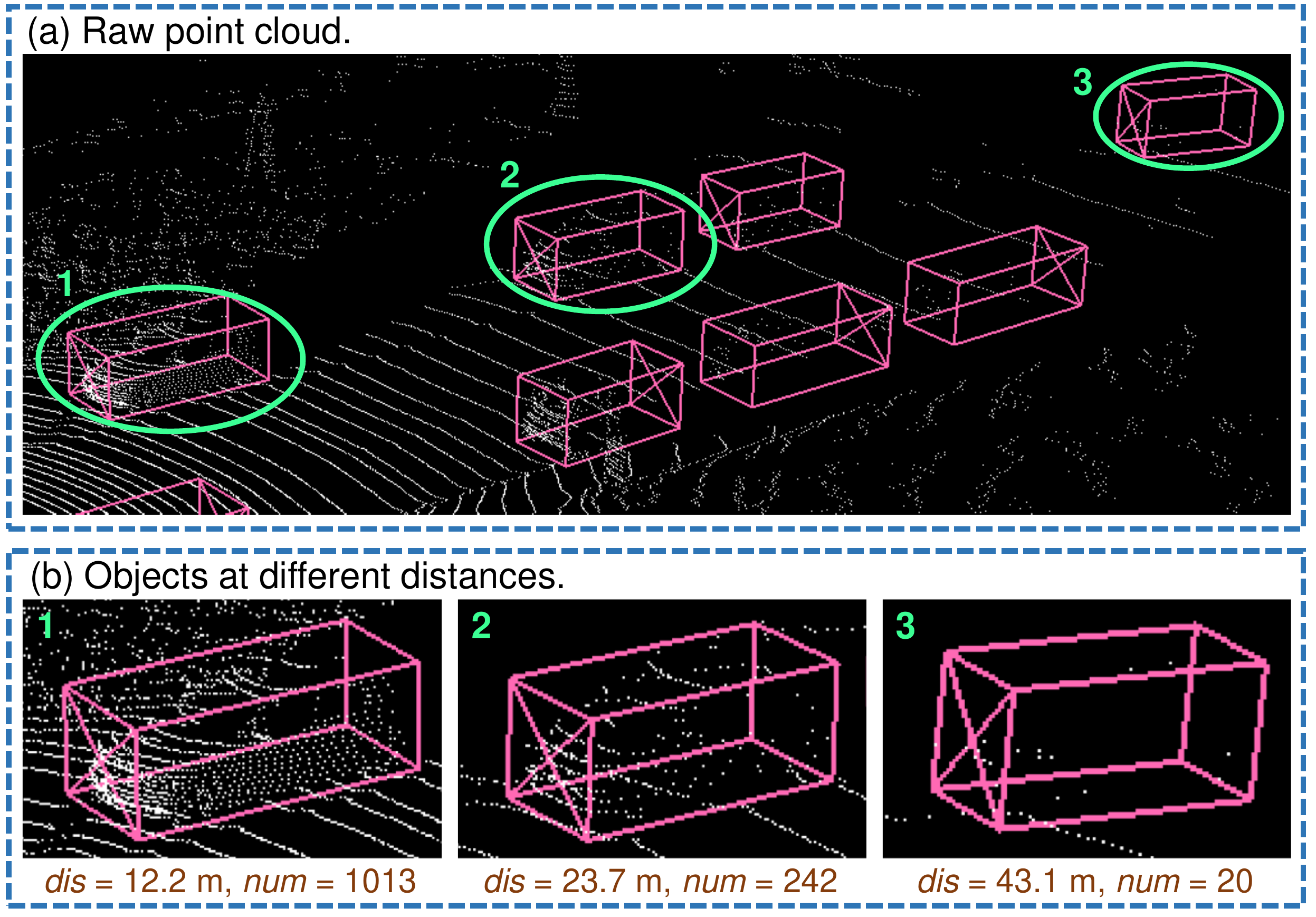}
	\caption{Illustration of the point cloud with imbalanced density on the KITTI dataset. (a) is a raw point cloud with annotated 3D bounding boxes labeled in pink. We zoom in three objects at different distances in (b), where $ dis $ and $ num $ denote the distance from the sensor and number of foreground points respectively.} \label{fig_imbalanced_density}
\end{figure}

Since object detection servers for the perception system of autonomous vehicles, the further away an object can be detected, the more time is left for the decision planning system, thus autonomous vehicles will be safer. As shown in Fig. \ref{fig_range_results_3d}, most of the previous works perform well at the near distance (AP \textgreater 90), but their performances drastically drop at the far distance (AP of 30$\sim$40). To some extent, it is more necessary to enhance the detection accuracy at far distances for safety concerns. For the large-scale point cloud scene, the density of points is imbalanced, \textit{i.e.}, the point cloud at the far-range area is more sparse. Nonetheless, most of the previous methods ignore this data-level issue. As demonstrated in Fig. \ref{fig_imbalanced_density}, the farther away the object is from the sensor, the fewer points are collected, the less spatial information is contained, restricting the feature representation and box prediction, thus it is indispensable to seek a solution to this imbalanced density problem.      

In this paper, we present a novel, unified and effective \textit{Spatial Information Enhancement Network} (SIENet), which is robust to the point clouds with unbalanced density and achieves accurate 3D object detection. Concretely, we design the \textit{Hybrid-Paradigm Region Proposal Network} (HP-RPN), which consists of the spconv branch for extracting multi-scale features from point clouds and generating proposals, the auxiliary branch for guiding the model to learn spatial information, and the keypoint branch for dynamically encoding and reweighting keypoint features from the imbalanced point cloud. In the second stage, we propose the \textit{Spatial Information Enhancement} (SIE) module, where the dense spatial shapes are predicted from the incomplete point sets within candidate boxes. Notably, the predicted shapes are uniform and dense, thus they contain more semantic cues to depict the proposals. Following that, the structure information is extracted and fused into the features from stage-1. Attributed to this mechanism, the features are enhanced with sufficient spatial information and can be used for refining more accurate 3D bounding boxes.     

Our primary contributions can be summarized into three-fold.

\begin{itemize} 	
	\item We design a novel spatial information enhancement module to predict the dense shapes of point sets in the candidate boxes, and learn the structure information to improve the ability of feature representation. Experimental results show that our proposed SIE module alleviates the suffering from point clouds with imbalanced density and improves the succeeding box refinement and confidence prediction.
	
	\item We present the hybrid-paradigm region proposal network for more effective multi-scale feature extraction and high-recall proposal generation from point clouds with imbalanced density.
	
	\item Our proposed SIENet achieves the state-of-the-art performance of 3D object detection on the KITTI benchmark. Besides, the encouraging experiment results show the outstanding improvement in far-range detection, as illustrated in Fig. \ref{fig_range_results_3d}.
	
\end{itemize}

\section{Related Works}

\subsection{3D Object Detection Based on Multi-modality Fusion}

Fusing the features from multiple sensors is the main idea of this series of work \cite{mv3d,avod,3dcvf,epnet,frustum,mmf}. MV3D \cite{mv3d} projects the point clouds into bird-eye view (BEV) to generate proposals, and fuses the features from different views to produce final predictions. AVOD \cite{avod} expands MV3D by exploiting the fusion of BEV feature maps and image feature maps to produce candidate boxes. Frustum PointNet \cite{frustum} first uses a mature 2D detector to generate candidate boxes on images, then extracts the point clouds from the 3D bounding frustums for box refinement. EPNet \cite{epnet} utilizes a two-stream architecture to learn pixel-wise and point-wise features respectively, and fuses the features from different modalities. 3D-CVF \cite{3dcvf} transforms the image features into the BEV domain and uses the attention mechanism to combine the image features and point cloud features. These methods require images as extra inputs, and the fusion step is very sensitive to data alignment and might introduce interference. Besides, the high time cost of projection also limits their inference efficiency. Without extra image data, our proposed framework directly uses raw point cloud as input to produce high-quality 3D bounding boxes.

\subsection{3D Object Detection Based on LiDAR Only}

According to the representations of point clouds, LiDAR-based 3D detectors can be divided into voxel-based \cite{part, second, voxelnet} and point-based \cite{pointrcnn, votenet, std, 3dssd} methods. For voxel-based methods, point clouds are voxelized before being sent to the network for learning features. In VoxelNet \cite{voxelnet}, points are equally divided into voxels and voxel feature encoding layers are proposed to encode the voxel-wise features, which are subsequently connected to RPN to predict 3D boxes. SECOND \cite{second} designs an effective sparse convolution algorithm, which can significantly increase the speed of training and inference. Part-$ A^2 $ \cite{part} proposes a successful two-stage framework to learn and group intra-object part information for accurate detection.

Point-based methods directly apply feature extraction modules, such as PointNet \cite{pointnet} and PointNet++ \cite{pointnet++}, on point clouds to learn global and local representations, and the back-end detection head generates high-quality 3D bounding boxes. PointRCNN \cite{pointrcnn} is a typical two-stage detector, where stage-1 learns point-wise features and generates candidate boxes, while stage-2 takes the output from stage-1 to refine the proposals. STD \cite{std} proposes an effective spherical anchor mechanism to reduce computation and design the IoU branch to improve the localization accuracy. 3DSSD \cite{3dssd} proposes a novel fusion sampling strategy to conserve the foreground points of various samples. Generally, point-based methods can learn more discriminative features but trigger higher computation cost, while voxel-based methods are more computationally efficient but inevitably lose information. Most of the previous voxel-based and point-based methods generally neglect the imbalanced density of point clouds, which is fatal for localization. In contrast, our SIENet is elaborately designed to address the density imbalance problem and further improve the detection performance.

\subsection{3D Shape Completion}

\begin{figure*}[htb]
	\centering	
	\includegraphics[width=\textwidth]{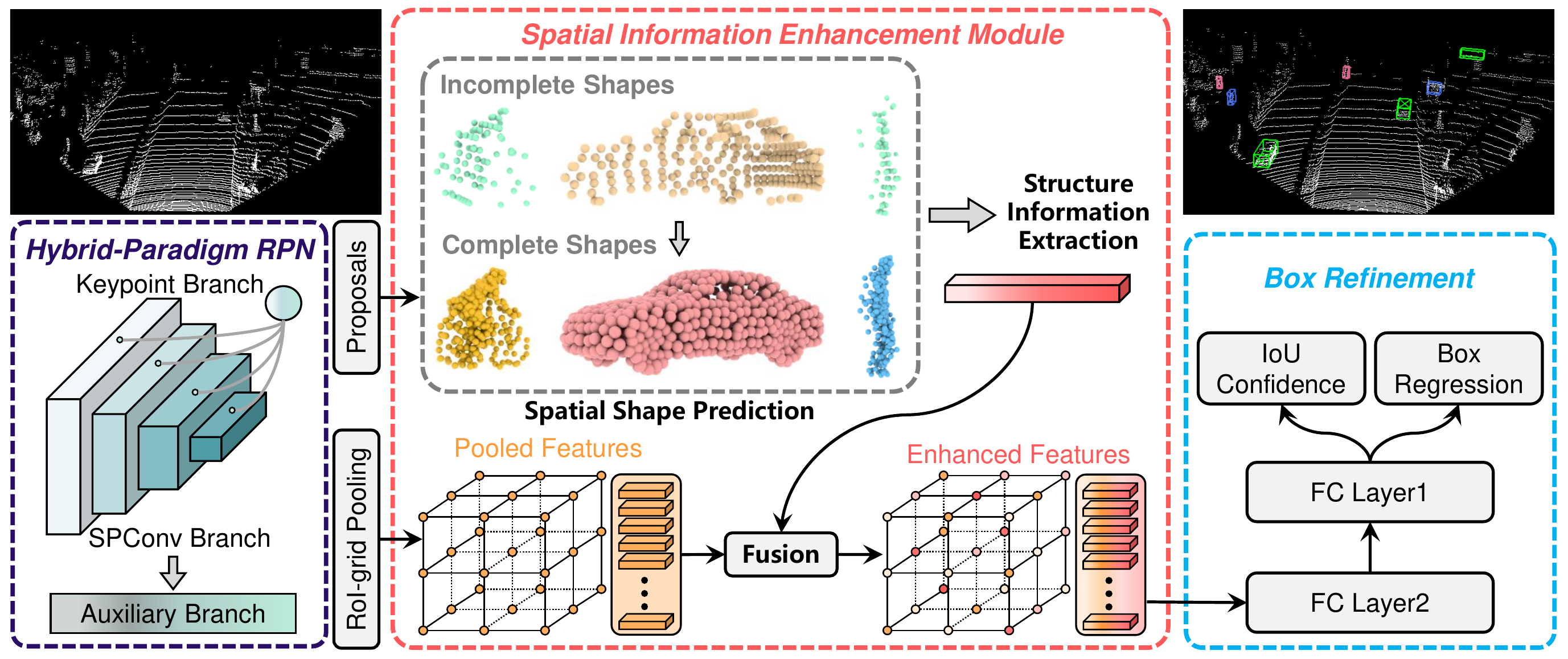}	
	\caption{The pipeline of our proposed Spatial Information Enhancement Network (SIENet). The whole framework consists of two stages: stage-1 uses the hybrid-paradigm RPN (HP-RPN) to extract features and generate accurate proposals, while stage-2 is designed to produce high-quality 3D bounding boxes via spatial information enhancement (SIE) module.}\label{framework}
\end{figure*}

There are several streams of approaches tackling 3D shape completion on point clouds. In \cite{lsm, robust, field}, the geometry-based inference is performed on incomplete input, while the interpolations are generated to recover the complete shape. In \cite{example, database}, the fragmentary input is first compared with several template models in the database, then the retrieved models are warped and consistently blended to obtain the 3D shape of objects. With the development of deep learning, many current works \cite{pcn, high, 3dcapsule, pfnet} leverage deep neural networks to create a mapping relationship from partial input to completed shape. Inspired by the 3D shape completion task, we design a novel mechanism to enhance the features via spatial information, and we formalize this mechanism as the SIE module, which first predicts the complete shapes of objects in candidate boxes, then extracts spatial semantic information to enhance the features from stage-1 for box refinement at stage-2.

\section{Our Framework}

In this section, we introduce the SIENet for 3D object detection from point clouds. Sec. \ref{stage1} introduces architecture details of our HP-RPN. Sec. \ref{stage2} presents the SIE module, a novel solution to the imbalanced density of point clouds. Sec. \ref{loss} discusses the loss functions. Our pipeline is displayed in Fig. \ref{framework}.

\subsection{Hybrid-Paradigm RPN}
\label{stage1}

\begin{figure}[t]
	\centering	
	\includegraphics[width=\linewidth]{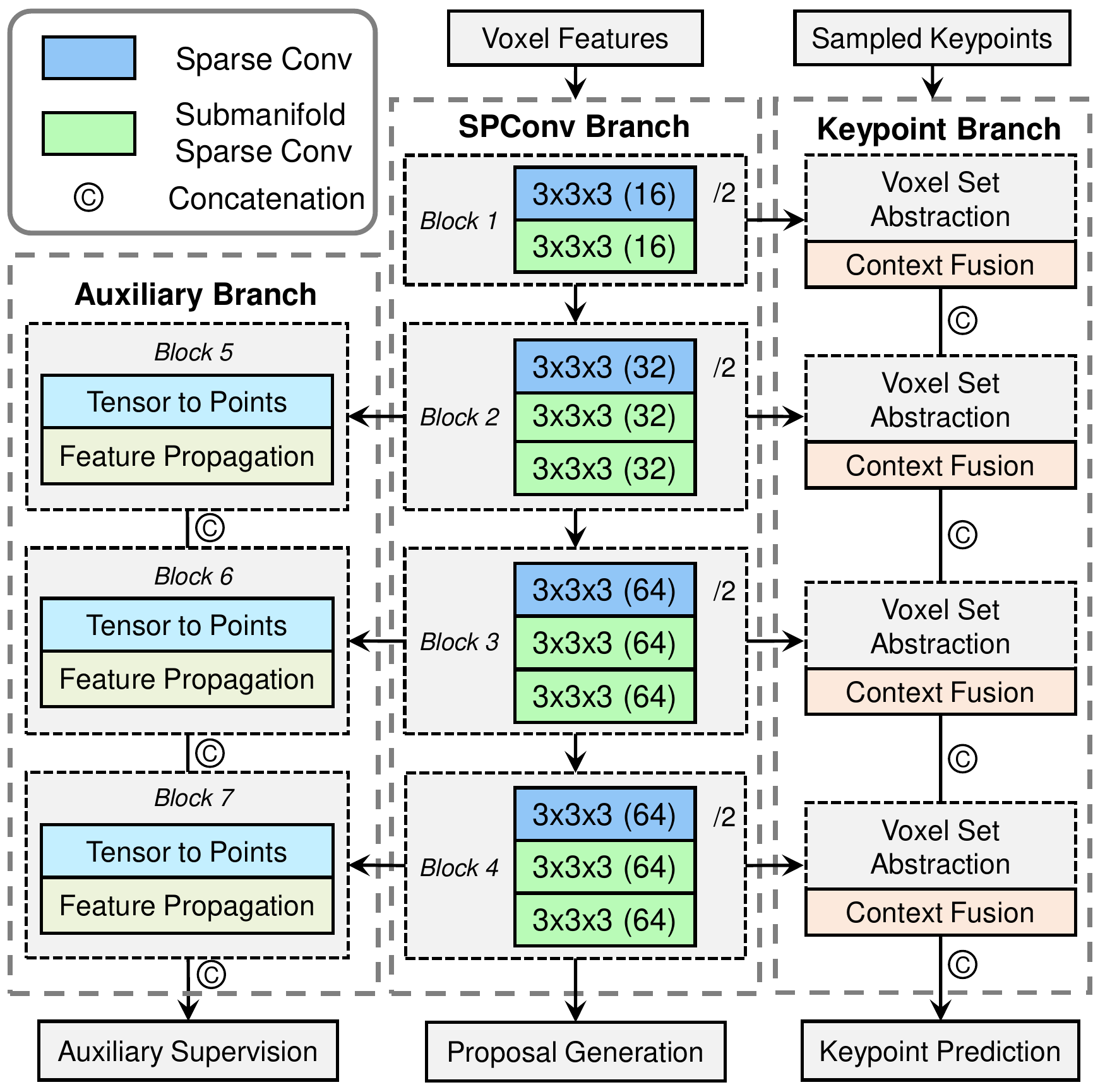}
	\caption{Network architecture of the HP-RPN. We use (kernel size) (channel) / (stride) to format the SPConv layers.}
	\label{architecture_hprpn}
\end{figure}

As shown in Fig. \ref{architecture_hprpn}, our HP-RPN consists of the SPConv branch, the auxiliary branch, and the keypoint branch. Specifically, we first voxelize the point clouds and use the spconv branch to learn more abstract voxel features. Concurrently, the keypoint branch utilizes the attention mechanism to dynamically encode the corresponding voxel features. Moreover, we adopt an auxiliary branch to guide network learning structure information.

\paragraph{SPConv branch.}

To efficiently learn the multi-scale features of point clouds, we first divide the point cloud into equally spaced voxels like \cite{voxelnet, second, fastpointrcnn}, and use the average coordinates and reflectances of the points within each voxel as the initial features. The SPConv branch stacks 4 sparse convolution blocks (block $ 1\sim4 $) to 8-times downsample the input voxel features. Each block uses one sparse convolution layer to reduce the size of feature maps in a computationally efficient fashion, followed by two submanifold sparse convolution layers for increasing the number of channels. These blocks process the voxel features layer by layer and produce discriminative features with smaller resolution. Following that, a general RPN head \cite{second, part} is expanded to generate high-recall candidate bounding boxes.

\paragraph{Auxiliary branch.}
Consecutive decreasing the resolution will lose the structure information of the objects with fewer points, which is fatal for RPN to generate proposals at the area with lower density of points. To tackle this obstacle, we append the auxiliary network \cite{sassd} to our first stage. Specifically, three upsampling blocks (block $ 5\sim7 $) respectively receive the outputs from the SPCconv branch, then convert the voxel coordinates into corresponding point coordinates in the real scene. Furthermore, the feature propagation layer \cite{pointnet++} is applied to interpolate the subsampled points. Finally, all the point-wise features are concatenated to learn the structure knowledge of objects. 

\paragraph{Keypoint branch.}

Voxel set abstraction \cite{pvrcnn} (VSA) is an effective module for encoding multi-scale information from voxel feature maps of different levels. Specifically, 2048 keypoints are selected by the furthest point sampling (FPS) algorithm, and each keypoint encodes its neighboring non-empty voxel-wise features. However, due to the density imbalance problem, not all keypoints make equally crucial contributions. Thus we expend the VSA layer with the context fusion layer \cite{deformable} to dynamically reweight the keypoint features. In our implementation, two multi-layer perceptron (MLP) modules are used to learn the point-wise attention map, which can suppress the interference keypoints and enhance the keypoints with more clear spatial semantics.  

\subsection{Spatial Information Enhancement Module}
\label{stage2}

For a candidate box generated from stage-1, the more dense the foreground point set is, the more spatial information can be conserved. Thus, the central idea of our SIE module is to predict the complete shapes and extract the structure information to improve the feature representation. To this end, we need to address three sub-tasks respectively, \textit{i.e.}, how to predict the spatial shapes, how to extract the structure information, and how to fuse it into the model for further box refinement. 

\paragraph{Spatial shape prediction.}

\begin{figure}[t]
	\centering
	\includegraphics[width=\linewidth]{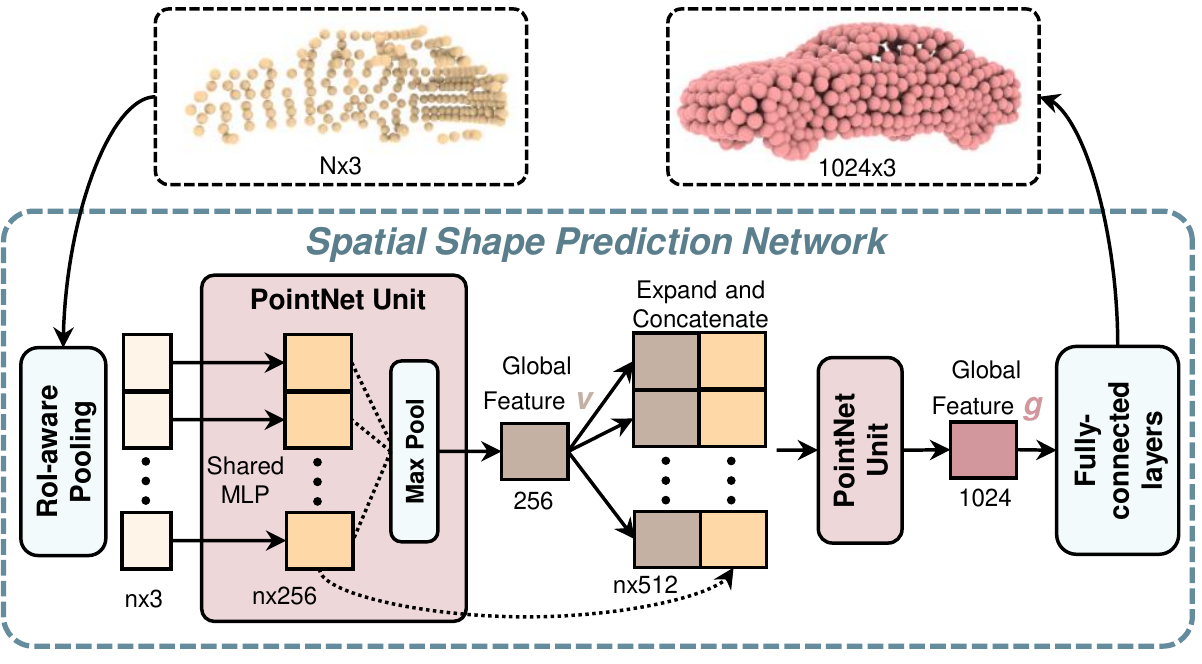}
	\caption{Illustration of the spatial shape prediction network. This network takes the incomplete point set as input and predicts the corresponding complete shape.} \label{architecture_pcn}
\end{figure}

The foreground points in a candidate box constitute a shape depicting the semantic cues, whereas, this shape is generally incomplete. Thus, we build upon a mature 3D shape completion framework PCN \cite{pcn} and design an effective spatial shape prediction network, which generates the dense shapes via completing the missing regions of the objects in proposals. As illustrated in Fig. \ref{architecture_pcn}, this network takes the incomplete point set (yellow points) as input and predicts the corresponding dense shape (pink points) in an encoder-decoder style. 

Given a candidate box, we can represent the inside point set as  $\left\{ \bm{P}_i|i=1,...,N\right\}$, where $\bm{P}_i$ is a vector of the coordinates $(x,y,z)$ and $N$ is the number of points. To reduce the interference information involved by inaccurate proposals, we first conduct RoI-aware pooling \cite{part} on the input point set $\bm{P}$ to get the output point set $\bm{P}'$ with a fixed size($r_p \times r_p \times r_p \times 3$), where $r_p$ is the pooling size. For simplicity, we adopt a $n\times 3$ matrix $\bm{M}$ to indicate the output of RoI-aware pooling, where $n = r_p^3$. Then a PointNet unit (MLP + Max Pooling) is adopted to capture the 256-dim global feature $\bm{v}$. We expand and concatenate $\bm{v}$ with the intermediate features of PointNet Unit, and use another PointNet unit to capture the generate the 1024-dim global feature $\bm{g}$. Finally, several fully-connected (FC) layers are stacked to generate $\hat{\bm{P}}$, which is a $1024\times3$ matrix that represents the dense and complete spatial shape. Different from the coarse-to-fine pipeline in PCN, we hold the view that the coarse output is sufficient for follow-up processing, thus we remove the detailed output branch, saving the GPU memory by the way. More training details and visualization results of the spatial shape prediction network are provided in the supplementary.

\paragraph{Structure information extraction.}
\label{Spatial Information Extraction}

To capture both the local and global context from the predicted shapes, we utilize a PointNet++ module to extract spatial information. Suppose that we have a fine-grained point set $\left\{\hat{\bm{P}_i}|i=1,...,1024\right\}$, we first use FPS algorithm to select $m$ points $\left\{\bm{S}_i|i=1,...,m\right\}$ from $\hat{\bm{P}}$. For each point in $\bm{S}$, we group its $T$ neighboring points from $\hat{\bm{P}}$, then we obtain a tensor of size $m\times T \times 3$. Finally, the PointNet units are applied to capture the local context of each point in $\bm{S}$, hence we get a matrix of size $m\times C_1$, where $ C_1 $ represents the number of channels. Additionally, the multi-scale grouping (MSG) strategy is also applied to capture the structure information. Specifically, we set two different radiuses in the grouping step and concatenate the output tensors at the channel dimension, \textit{i.e.}, the tensor of size $m\times (C_1+C_1)$, which is then reshaped and transferred into a FC layer to reduce the number of channels and generate the global structure information $ \bm{F}^{s} \in \mathbb{R}^{C_1} $. 

\paragraph{Feature fusion and box refinement.}

As exhibited in Fig. \ref{framework}, given a series of proposals and keypoints features learned from stage-1, we use RoI-grid pooling \cite{pvrcnn} to capture gird point context information among the neighboring keypoints features. Concretely, the $6 \times 6 \times 6$ grid point features are represented as $\bm{F}^{g}=\left\{\bm{f}^{g}_1,\bm{f}^{g}_2,...,\bm{f}^{g}_{216} \right\} \in \mathbb{R}^{216\times C_2}$, where $ C_2 $ represents the number of channels and $ \bm{f}^{g}_i $ is the feature vector of $ i $-th grid point. Subsequently, the structure information $ \bm{F}^{s}$ is expanded and concatenated with grid point features $ \bm{F}^{g} $ to form the concatenation feature $ \bm{F}^{c}=[\bm{F}^{g}, \bm{F}^{s}] \in \mathbb{R}^{216\times (C_2+C_1)} $. 

However, this simplest fusion strategy neglects the global dependencies within proposals. Besides, $ \bm{F}^{s} $ and $ \bm{F}^{g} $ come from different modalities, concatenating them without any extra processing might introduce interference. Therefore, we adopt the perspective-channel attention \cite{3diounet} to dynamically reweight the concatenated feature $ \bm{F}^{c} $ and obtain the enhanced feature $ \bm{F}^{e} \in \mathbb{R}^{216\times (C_2+C_1)} $. After that, two branches of MLP are used for final box refinement and confidence prediction respectively. More details of the feature fusion are described in the supplementary.

\subsection{Loss Function}
\label{loss}

% Table of both car and cyclist.
\begin{table*}
	
	\resizebox{\textwidth}{!}{
		\begin{tabular}{c|c|c||
				p{0.85cm}<{\centering} p{0.85cm}<{\centering} p{0.85cm}<{\centering}|
				p{0.85cm}<{\centering} p{0.85cm}<{\centering} p{0.85cm}<{\centering}|
				p{0.85cm}<{\centering} p{0.85cm}<{\centering} p{0.85cm}<{\centering}|
				p{0.85cm}<{\centering} p{0.85cm}<{\centering} p{0.85cm}<{\centering}
			}
			\hline
			\multirow{2}{*}{Method} &
			\multirow{2}{*}{Reference} &
			\multirow{2}{*}{Modality} &
			\multicolumn{3}{c|}{Car - 3D Detection} &
			\multicolumn{3}{c|}{Car - BEV Detection} &
			\multicolumn{3}{c|}{Cyclist - 3D Detection} & 
			\multicolumn{3}{c}{Cyclist - BEV Detection}\\
			
			\cline{4-15}
			& & & Easy & \underline{Mod.}  &Hard & Easy & \underline{Mod.} & Hard & Easy & \underline{Mod.} & Hard & Easy & \underline{Mod.} & Hard \\
			
			\hline
			\hline
			MV3D \cite{mv3d} & CVPR 2017 & RGB + LiDAR & 
			74.97 & 63.63 & 54.00 & 86.62 & 78.93 & 69.80 & -     & -     & -  & -     & -     & -   \\
			AVOD \cite{avod} & IROS 2018 & RGB + LiDAR & 
			76.39 & 66.47 & 60.23 & 89.75 & 84.95 & 78.32 & 57.19 & 42.08 & 38.29 & 64.11 & 48.15 & 42.37\\
			
			Frustum PointNet \cite{frustum}  & CVPR 2018 & RGB + LiDAR & 82.19 & 69.79 & 60.59 & 91.17 & 84.67 & 74.77 & 72.27 & 56.12 & 49.01 & 77.26 & 61.37 & 53.78
			\\
			
			UberATG-ContFuse \cite{contfuse}  & ECCV 2018 & RGB + LiDAR & 83.68 & 68.78 & 61.67 & 94.07 & 85.35 & 75.88 & - & - & - & - & - & - \\
			
			UberATG-MMF \cite{mmf}  & CVPR 2019 & RGB + LiDAR & 88.40 & 77.43 & 70.22 & 93.67 & 88.21 & 81.99 & - & - & - & - & - & - \\

			EPNet \cite{epnet}  & ECCV 2020 & RGB + LiDAR & 89.81 & 79.28 & 74.59 & 94.22 & 88.47 & 83.69 & - & - & - & - & - & - \\
			
			3D-CVF \cite{3dcvf}  & ECCV 2020 & RGB + LiDAR & 89.20 & 80.05 & 73.11 & 93.52 & 89.56 & 82.45 & - & - & - & - & - &-\\

			\hline
			
			SECOND \cite{second} & Sensors 2018 & LiDAR only & 83.34 & 72.55 & 65.82 & 89.39 & 83.77 & 78.59 & 71.33 & 52.08 & 45.83 & 76.50 & 56.05 & 49.45\\
			
			PointPillars \cite{pointpillars} & CVPR 2019 & LiDAR only & 82.58 & 74.31 & 68.99 & 90.07 & 86.56 & 82.81 & 77.10 & 58.65 & 51.92 & 79.90 & 62.73 & 55.58\\
			
			PointRCNN \cite{pointrcnn} & CVPR 2019 & LiDAR only & 86.96 & 75.64 & 70.70 & 92.13 & 87.39 & 82.72 & 74.96 & 58.82 & 52.53 & 82.56 & 67.24 & 60.28\\
			
			Fast PointRCNN \cite{fastpointrcnn} & ICCV 2019 & LiDAR only & 85.29 & 77.40 & 70.24 & 90.87 & 87.84 & 80.52 & - & - & - & - & - & -\\
			
			STD \cite{std} & ICCV 2019 & LiDAR only & 87.95 & 79.71 & 75.09 & 94.74 & 89.19 & \textbf{86.42} & 78.69 & 61.59 & 55.30 & 81.36 & 67.23 & 59.35\\
			
			Part-$ A^2 $ \cite{part} & TPAMI 2020 & LiDAR only & 87.81 & 78.49 & 73.51 & 91.70 & 87.79 & 84.61 & 79.17 & 63.52 & 56.93 & 83.43 & 68.73 & 61.85\\

			Associate-3Ddet \cite{associate} & CVPR 2020 & LiDAR only & 85.99 & 77.40 & 70.53 & 91.40 & 88.09 & 82.96 & - & - & - & - & - & -\\
			
			SERCNN \cite{sercnn} & CVPR 2020 & LiDAR only & 87.74 & 78.96 & 74.30 & 94.11 & 88.10 & 83.43 & - & - & - & - & - & -\\
			
			3DSSD \cite{3dssd} & CVPR 2020 & LiDAR only & 88.36 & 79.57 & 74.55 & 92.66 & 89.02 & 85.86 & 82.48 & 64.10 & 56.90 & \textbf{85.04} & 67.62 & 61.14\\
			
			PV-RCNN \cite{pvrcnn} & CVPR 2020 & LiDAR only & \textbf{90.25} & 81.43 & 76.82 & 94.98 & 90.65 & 86.14 & 78.60 & 63.71 & 57.65 & 82.49 & 68.89 & 62.41\\
			
			SA-SSD \cite{sassd} & CVPR 2020 & LiDAR only & 88.75 & 79.79 & 74.16 & \textbf{95.03} & \textbf{91.03} & 85.96 & - & - & - & - & - & -\\
			
			Point-GNN \cite{pointgnn} & CVPR 2020 & LiDAR only & 88.33 & 79.47 & 72.29 & 93.11 & 89.17 & 83.90 & 78.60 & 63.48 & 57.08 & 81.17 & 67.28 & 59.67\\
			
			\hline
			SIENet (Ours) & - & LiDAR only & 88.22 & \textbf{81.71} & \textbf{77.22} & 92.38 & 88.65 & 86.03 & \textbf{83.00} & \textbf{67.61} & \textbf{60.09} & 84.64 & \textbf{71.21} & \textbf{64.61} \\
			
			\hline
	\end{tabular}}
	\caption{Comparison with the state-of-the-art methods on KITTI test set. The official ranking metric of KITTI online server is moderate AP with 40 recall positions. We can see that our SIENet achieves the top performance on the more challenging cases, \textit{i.e.}, highest moderate and hard AP for both car detection and cyclist detection.}\label{testset}
	
\end{table*}

The proposed SIENet is trained in end-to-end manner. Our overall loss includes $L_{rpn}$ of HP-RPN in stage-1 and box refinement loss $L_{rcnn}$ in stage-2 as: 

\begin{equation}
L = L_{rpn} + L_{rcnn}.
\end{equation}

The HP-RPN loss $L_{rpn}$ is defined as the summation of box generation, keypoint prediction and auxiliary supervision as:

\begin{equation}
L_{rpn} = L_{box} + L_{pt} + L_{aux},
\end{equation} where the box generation loss $L_{box}$ and keypoint prediction loss $L_{pt}$ are same as \cite{pvrcnn}. We set the foreground segmentation and offset regression as the learning task of our auxiliary supervision: 

\begin{equation}
L_{aux} = L_{seg} + L_{offset}. 
\end{equation} 

For the foreground segmentation, we need to segment the positive voxels which belong to ground-truth boxes, but a great majority of them are negative samples. To better learn from this serious imbalanced data, we adopt the focal loss \cite{focalloss} to distinguish whether the voxels belong to positive samples, thus the segmentation loss $L_{seg}$ is formulated as follows: 

\begin{equation}
\label{focal_loss}
L_{seg} = \alpha_t (1-p_t)^{\gamma}log(p_t),
\end{equation}where $p_t$ is the predicted possibility of positive samples, $\alpha$ and $\gamma$ are the constants to control the weighting of sampling balance, we keep the default parameters $\alpha_t=0.25$ and $\gamma=2$. The offset regression loss $ L_{offset} $ is a binary cross entropy loss defined as follows:

\begin{equation}
L_{offset} = -u^tlog(u^p)-(1-u^t)log(1-u^p), 
\end{equation} where $ u^t $ is the target of regressing offsets between [0,1], and $ u^p $ is the predicted offset value calculated by sigmoid activate function. We adopt the same box refinement loss $ L_{rcnn} $ of stage-2 as \cite{pvrcnn}.

\section{Experiments}

We evaluate our proposed SIENet on KITTI \cite{kitti}, which is the most popular 3D object detection benchmark. KITTI provides 7,481 samples for training and 7,518 samples for testing. Following the frequently used \textit{train}/\textit{val} split, we divide the training samples into \textit{train} split of 3,712 samples and \textit{val} split of 3,769 samples. The online benchmark of KITTI adopts average precision (AP) with a 3D overlap threshold of 0.7 as the evaluation metric of the car class (0.5 for pedestrian and cyclist), where three difficulty levels (easy, moderate and hard) are taken into consideration. We compare the SIENet with previous state-of-the-art methods on both \textit{val} split and test set.

\subsection{Implementation Details}
\label{implementation details}

\paragraph{Training and inference settings.}

\begin{table}
	\centering
	\resizebox{8cm}{!}{
		
		\begin{tabular}
			{c|c|c|c}
			
			\hline
			Method & Reference & Modality & 3D AP\\
			
			\hline
			
			MV3D \cite{mv3d} & CVPR 2017 & RGB + LiDAR & 62.68  \\
			
			AVOD \cite{avod} & IROS 2018 & RGB + LiDAR & 74.44 \\
			
			Frustum PointNet \cite{frustum} & CVPR 2018 & RGB + LiDAR & 70.92 \\
			
			3D-CVF \cite{3dcvf} & ECCV 2020  & RGB + LiDAR & 79.88  \\
			
			\hline
			
			SECOND \cite{second} & Sensors 2018 & LiDAR only & 76.48  \\
			
			PointRCNN \cite{pointrcnn} & CVPR 2019 & LiDAR only & 78.63  \\
			
			Fast PointRCNN \cite{fastpointrcnn} & ICCV 2019 & LiDAR only & 79.00 \\
			
			STD \cite{std} & ICCV 2019 & LiDAR only & 79.8 \\
			
			Part-$ A^2 $ \cite{part} & TPAMI 2020 & LiDAR only & 79.47 \\
			
			Associate-3Ddet \cite{associate} & CVPR 2020 & LiDAR only & 79.17\\
			
			SERCNN \cite{sercnn} & CVPR 2020 & LiDAR only & 79.21 \\
			
			3DSSD \cite{3dssd} & CVPR 2020 & LiDAR only & 79.45 \\
			
			SA-SSD \cite{sassd} & CVPR 2020 & LiDAR only & 79.91 \\
			
			Point-GNN \cite{pointgnn} & CVPR 2020 & LiDAR only & 78.34 \\
			
			PV-RCNN \cite{pvrcnn} & CVPR 2020 & LiDAR only & 83.90 \\
			
			\hline
			
			SIENet (Ours) & - & LiDAR only & \textbf{84.40} \\
			
			\hline		
	\end{tabular}}
	\caption{Comparison with the state-of-the-art methods on KITTI \textit{val} split, where the moderate AP for car class is calculated with 11 recall positions.}\label{valset}
	
\end{table}

To relief the burden of training, we train the spatial shape prediction network of SIE module in advance, and keep the parameters frozen while training. We provide the details of dataset pre-processing and parameter optimizing in the supplementary. We train the SIENet on 4 NVIDIA TITAN XP GPUs for 80 epochs, which takes around 13 hours. Each mini-batch consists of 8 training samples evenly distributed on 4 GPUs. We apply ADAM \cite{adam} optimizer and the initial learning rate of 0.01 with cosine annealing strategy. In the training stage, we assign 128 proposals for each scene and 50\% of them are positive proposals, \textit{i.e.}, which have 3D IoU more than the threshold of 0.55 with corresponding ground truth boxes.

When in the inference stage, non-maximum suppression (NMS) is conducted with a threshold of 0.7 to filter the redundant proposals, and only the top-100 proposals are kept for the follow-up refinement. Finally, another NMS with a threshold of 0.01 is used to remove the redundant box predictions.

\paragraph{Data augmentation.}

% 1 paragraph of augmentation details
Without bells and whistles, we only apply several commonly used augmentation methods. Specifically, we randomly flip each scene along \textit{X}-axis with 50\% probability, and then rotate it around \textit{Z}-axis with a random angle sampled from $ \left[ -\frac{\pi }{4}, \frac{\pi }{4} \right] $, finally we scale the point cloud with a uniformly sampled factor from $ \left[ 0.95, 1.05 \right] $. GT sampling augmentation proposed by \cite{second} is adopted in all recent state-of-the-art LiDAR-based 3D object detection frameworks, we utilize this augmentation strategy and set the maximum of samplers for each scene to 15.  

\begin{figure*}[t]
	\centering	
	\includegraphics[width=\textwidth]{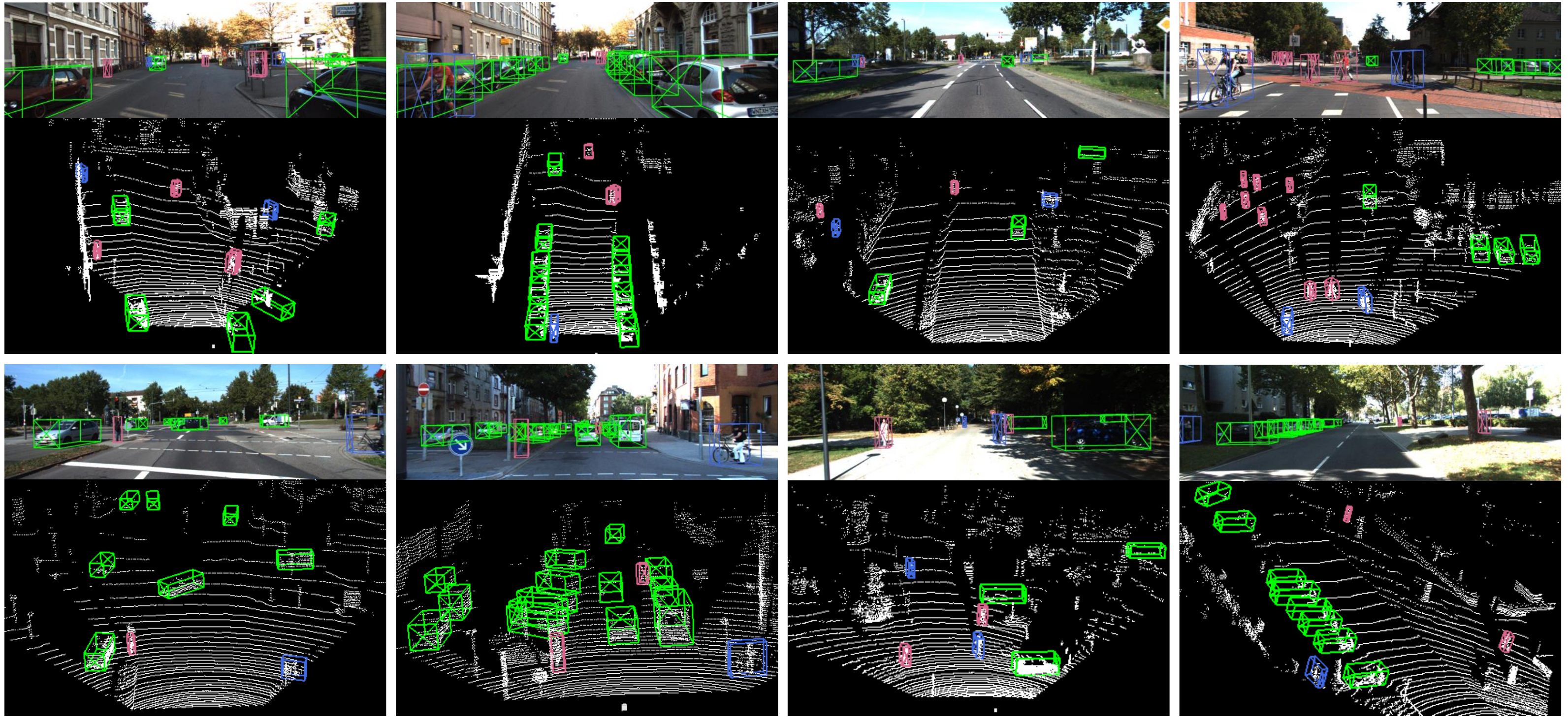}
	\caption{Qualitative results of SIENet on KITTI dataset. The detection results are demonstrated on images (upper) and the corresponding point clouds (lower). For each box, we use the X to specify the driving direction. The predicted bounding boxes of cars, cyclists, and pedestrians are drawn in green, pink, and blue, respectively.}
	\label{vis_results}
\end{figure*}

\begin{table}
	\resizebox{\linewidth}{!}{
		\begin{tabular}
			{c|p{0.85cm}<{\centering} p{0.85cm}<{\centering} p{0.85cm}<{\centering}|p{0.85cm}<{\centering} p{0.85cm}<{\centering} p{0.85cm}<{\centering}|c}
			
			\hline
			IoU &
			\multicolumn{3}{c|}{3D Detection } & 
			\multicolumn{3}{c|}{BEV Detection} & FPS\\
			
			\cline{2-7}
			Thresh. & Easy & Mod. & Hard & Easy & Mod. & Hard & (Hz)\\
			
			\hline
			0.7 & 92.49 & 85.43 & 83.05 & 90.29 & 88.41 & 87.77 & 6.2\\

			\hline		
	\end{tabular}}
	\caption{Performance of car class on KITTI \textit{val} split with 40 recall positions.}
	\label{tab_val_40_recall}	
\end{table}

\subsection{Comparison with State-of-the-art Methods}

In this section, we conduct the comparisons with state-of-the-art methods on KITTI dataset. For the evaluation on test set, we use all 7,481 samples to train the model and submit the results to KITTI official server. For the comparison on \textit{val} split, we only train the model on \textit{train} split of 3712. 

\paragraph{Online evaluation on test set.}
We show the comparison with the state-of-the-art methods on KITTI test set in Table \ref{testset}, the results are obtained from the official leaderboard as of Mar. 10th, 2021. It is noted that our SIENet attains top performances on the 3D detection of both car and cyclist \footnote{All methods are ranked based on the moderate AP of KITTI official leaderboard.}. Our method outperforms all multi-modality fusion-based methods, including UberATG-MMF \cite{mmf}, EPNet \cite{epnet}, and 3D-CVF \cite{3dcvf}, by a very large margin (2.63 to 7 points of hard AP, and 1.66 to 4.28 points of moderate AP). Compared with the LiDAR-based methods, we also outperform the recent state-of-the-art detectors, \textit{e.g.}, Point-GNN \cite{pointgnn}, 3DSSD \cite{3dssd}, SA-SSD \cite{sassd}, and PV-RCNN \cite{pvrcnn} by 0.4 to 4.93 points of hard AP and achieve the highest moderate AP on car class. It is noted that SIENet achieves the highest performances under almost all metrics of cyclist detection with large improvements (3.51 points for moderate AP and 2.44 points for hard), which further depicts the superior generalization of our method.

\paragraph{Offline evaluation on val split.}
We demonstrate the 3D detection AP and BEV detection AP of our SIENet on KITTI \textit{val} split in Table \ref{tab_val_40_recall}, where the AP for car class is under the IoU threshold of 0.7 and calculated with 40 recall positions. To make the experimental results more convincing, we conduct a fair comparison between our SIENet and previous state-of-the-art detectors in Table \ref{valset}, where AP with 11 recall positions is adopted to evaluate the accuracy of car detection. Table \ref{valset} delivers that our SIENet outperforms all previous methods on KITTI \textit{val} split. Compared with the recent state-of-the-art detectors, \textit{e.g.}, Part-$ A^2 $, 3DSSD, SASSD, Point-GNN, and PV-RCNN, we obtain the 0.5 to 4.93 points of moderate AP gains. Besides, we measure the runtime of SIENet on an Intel i7-6700K CPU and a single NVIDIA TITAN XP GPU. As presented in Table \ref{tab_val_40_recall}, our method runs at 6.2 FPS, while PV-RCNN achieves 7.6 FPS with our same platform and test program, which means SIENet attains the top performance with only about 1 FPS of extra runtime cost. The visualizations of SIENet on the KITTI are illustrated in Fig. \ref{vis_results}, showing that our method can produce accurate 3D bounding boxes from the point clouds with imbalanced density.

\paragraph{Does the SIENet indeed alleviate the suffering from density imbalance problem?}

\begin{table}
	\resizebox{\linewidth}{!}{
		\begin{tabular}
			{c ||c c c | c | c p{1.1cm}<{\centering} c | c}
			
			\hline
			\multirow{2}{*}{Method} & 
			\multicolumn{4}{c|}{3D Detection } & 
			\multicolumn{4}{c}{BEV Detection } \\
			\cline{2-9}
			& 0-20m & 20-40m & 40m-Inf & Mean & 0-20m & 20-40m & 40m-Inf & Mean \\
			
			\hline
			SECOND \cite{second}& 89.99 & 76.13 & 38.73 & 68.28 & 90.49 & 86.80 & 55.68 & 77.66 \\
			
			PointPillars \cite{pointpillars}& 89.60 & 73.28 & 34.14 & 65.67 & 90.24 & 86.08 & 51.16 & 75.83\\
			
			Part-$ A^2 $ \cite{part}& \textbf{90.42} & 76.98 & 39.72 & 69.04 & \textbf{90.67} & 86.37 & 56.10 & 77.71\\
			
			PV-RCNN \cite{pvrcnn}& 90.29 & 76.83 & 42.15 & 69.76 & 90.48 & 86.36 & 57.44 & 78.09\\		
			\hline			
			
			SIENet (ours) & 90.33 & \textbf{82.11} & \textbf{46.54} & \textbf{72.99} & 90.60 & \textbf{87.00} & \textbf{59.70} & \textbf{79.10}\\
			
			\rowcolor{MyCyan}\textit{Improvement} & \textit{-0.09} & \textit{+5.13} & \textit{+4.39} & \textit{+3.23} & \textit{-0.07} & \textit{+0.20} & \textit{+2.26} & \textit{+1.01}\\
			\hline	
			
	\end{tabular}}
	\caption{Performance comparison at different distance ranges on the moderate level car class of KITTI \textit{val} split set. The results are evaluated with the moderate AP calculated by 11 recall positions.}\label{range_results_table}
\end{table}

As previously discussed, this density imbalance problem is a data-level issue related to objects' different distances to the sensor. Therefore, the performance at different distances is competent to be a credible benchmark for evaluating the robustness to imbalanced density. Concretely, we first download the pre-trained models from OpenPCDet\footnote{\url{https://github.com/open-mmlab/OpenPCDet}}, which is a successful open-source framework supporting many 3D detectors. Then we divide the KITTI \textit{val} split into three sub-sets according to the distance ranges. Finally, we evaluate the pre-trained models on these sub-sets. As shown in Table \ref{range_results_table}, our method achieves an average gain of 3.23 points for 3D AP among all distance ranges. Besides, our method achieves competitive AP at near distance, while with the distance increases, our method outperforms previous methods with 5.13 points AP gain at medium distance and 4.39 points AP gain at far distance. Note that the same trend also appears in BEV detection. We visualize the results of 3D AP in Fig. \ref{fig_range_results_3d}, revealing the consistent expectation that SIENet alleviates the suffering from point clouds with imbalanced density and boosts the detection performance at medium and far distances. It is noted that our model achieves top results on far object detection, the AP at near distance is slightly lower than some methods, as delivered in the $ 2^{nd} $ and $ 6^{th} $ columns of Table \ref{range_results_table}. We argue that this inconsistency is acceptable, because it is not hard for recent methods to accurately localize near objects with sufficient points, the factor that affects the overall accuracy is the performance at medium and long distances. Besides, for an autonomous vehicle at high speed, the farther the objects can be localized, the safer it will be. More quantitative results are provided in the supplementary.

\subsection{Ablation Studies}
In this section, we conduct a series of ablation studies to discuss our design choices of the proposed components. Following the general principles, all models are trained on KITTI \textit{train} split and evaluated on \textit{val} split.

%% R40
\begin{table}
	\centering
	\resizebox{0.9\linewidth}{!}{
		
		\begin{tabular}
			{p{1.2cm}<{\centering} | p{1.2cm}<{\centering} | p{1.2cm}<{\centering} ||p{0.8cm}<{\centering} p{0.8cm}<{\centering} p{0.8cm}<{\centering}}
			
			\hline
			RPN & Auxiliary & Context & \multicolumn{3}{c}{3D Detection (R40)} \\
			
			\cline{4-6}
			Baseline & Branch & Fusion & Easy & \underline{Mod.} & Hard \\
			\hline

			$ \checkmark $ & $ \times $ & $ \times $ & 91.95 & 84.48 & 82.67 \\
			
			$ \checkmark $ & $ \checkmark $ & $ \times $ & 92.28 & 84.82 & 82.75 \\

			$ \checkmark $ & $ \times $ & $ \checkmark $ & 92.21 & 84.83 & 82.69 \\

			$ \checkmark $ & $ \checkmark $ & $ \checkmark $ & \textbf{92.49} & \textbf{85.43} & \textbf{83.05}  \\
			\hline
			
	\end{tabular}}
	\caption{Effects of auxiliary branch and context fusion for HP-RPN. R40: 40 recall positions are used to calculate the results.}\label{ab_rpn}
	
\end{table}

\begin{table}
	\centering
	\resizebox{0.8\linewidth}{!}{
		\begin{tabular}
			{p{3.0cm}<{\centering} ||p{0.8cm}<{\centering} p{0.8cm}<{\centering} p{0.8cm}<{\centering}}
			
			\hline
			\multirow{2}{*}{Method} & 
			\multicolumn{3}{c}{3D Detection (R40)} \\
			
			\cline{2-4} & Easy & \underline{Mod.} & Hard \\
			\hline
			
			RCNN baseline & 91.97 & 84.59 & 82.63 \\
			
			\hline
			+ SIE-Con & 92.37 & 85.07 & 82.80 \\

			+ SIE-Att & \textbf{92.49} & \textbf{85.43} & \textbf{83.05} \\
			
			\hline		
	\end{tabular}}
	\caption{Effects of SIE module. SIE-Con: SIE module with concatenation. SIE-Att: SIE module with attention-based fusion.}\label{ab_rcnn}
	
\end{table}

\paragraph{Effects of HP-RPN.}

In Table \ref{ab_rpn}, we investigate the effectiveness of different structures in the HP-RPN. We remove the auxiliary branch and the context fusion layer in stage-1, and represent the remaining part as the baseline of RPN. As shown in the $ 1^{st} $ row, the performance drops a lot if we only keep the RPN baseline. The $ 2^{nd} $ and $ 3^{rd} $ rows show that both these two components outperform the independent RPN baseline. Note that in the last row, the combination of auxiliary branch and context fusion layer contributes an improvement of (0.54\%, 0.95\%, 0.38\%), validating the effectiveness of our HP-RPN.

\paragraph{Effects of SIE module.}

We explore the effects of SIE module in table \ref{ab_rcnn}. In the $ 1^{st} $ row, we remove the SIE module, which leads to the drastically drops of performance among all difficulty levels. Note that in the $ 2^{nd} $ row, we add the SIE module and just use the simplest concatenation fusion, which leads to a 0.4\% AP gain on the easy difficulty, and increases the moderate AP by 0.48\%. This improvement benefits from that the SIE module facilitates the model to learn better spatial information. Furthermore, We compare the effects of the attention-based fusion in the $ 3^{rd} $ row, where the total AP gains of (0.52\%, 0.84\%, 0.42\%) are achieved, demonstrating that the attention-based fusion can better capture and aggregate the features.

\section{Conclusion}
In this paper, we propose the novel SIENet for 3D object detection from point clouds with the imbalanced density. Concretely, we design the spatial information enhancement module to produce the complete shapes within proposals, and learn structure information to enhance the features for box refinement. Besides, we introduce the hybrid-paradigm RPN for effective feature extraction and proposal generation. Extensive experiments have validated the effectiveness of our proposed framework. 

{\small
\bibliographystyle{ieee_fullname}
\bibliography{ref}
}

\appendix
\section{Overview}

This document provides additional technical details, extra experimental results and more qualitative visualizations of the submitted paper. 

In Sec. \ref{sie}, we discuss the training details of the spatial shape prediction network, which is the essential component of our SIE module, and show the visualizations of the predicted shapes. In Sec. \ref{section_atfusion}, we describe the network architecture of another crucial element, the feature fusion network. In Sec. \ref{kitti}, we present more quantitative results on KITTI \cite{kitti}.

\section{Details of the Spatial Shape Prediction Network (Sec 3.2)}
\label{sie}

\subsection{Training Settings}
Spatial shape prediction is essential to our elaborately designed SIE module. To reduce the burden of training, we train our spatial shape prediction network in advance. Although current 3D shape completion methods can recover the shapes of commonly used categories, \textit{e.g.}, chair, table, piano, airplane and car, they may fail to complete the shapes of pedestrian and cyclist because no corresponding data can be obtained. Therefore, we extract the foreground points of pedestrians and cyclists in the KITTI train set, and filter the instances with few points. We use these samples as ground truth and randomly remove some points of them as incomplete input. For the data of the car class, we download the corresponding parts from ShapeNet \cite{shapenet}.

After solving the challenge of data, we train a model based on PCN \cite{pcn}. Concretely, we use Adam \cite{adam} optimizer with a starting learning rate at 0.0001, which decays by 0.7 in every 50,000 iterations. We train the model on one NVIDIA TITAN XP GPU with batch size 32 for 300,000 iterations. Furthermore, due to the limitation of GPU memory, we remove the detailed output branch after the model is trained. Finally, we initialize our spatial shape prediction network with the saved weights.

\subsection{Visualizations of the Predicted Shapes}

\begin{figure*}[h]
	\centering	
	\includegraphics[width=0.9\linewidth]{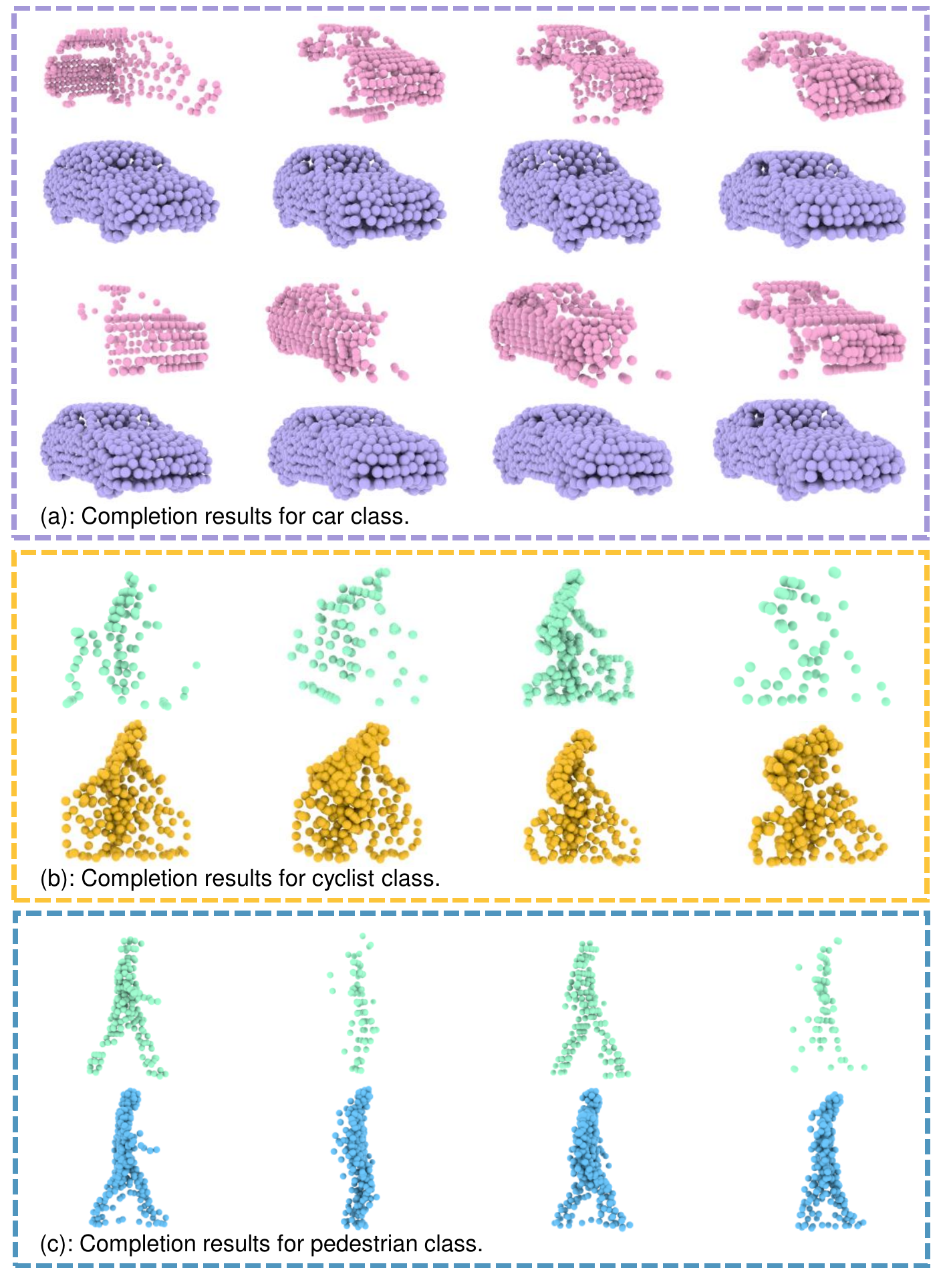}
	\caption{Visualizations of the predicted shapes within the proposals. For each candidate object, the upper part is the incomplete point set, while the lower part is the corresponding prediction.}
	\label{sip_results}
\end{figure*}

We visualize more results of our spatial shape prediction network, as shown in Fig. \ref{sip_results}, for each candidate object, the upper part is the incomplete point set, while the lower part is the corresponding prediction. From this figure, it can be seen that the input point sets are sparse and just exist on a fractional surface of the objects, while the output spatial shapes are dense and contain more semantic cues. Besides, our spatial shape prediction network performs well on all these three classes (car, cyclist, and pedestrian), demonstrating the promising generalization.

\begin{figure}[t]
	\centering	
	\includegraphics[width=\linewidth]{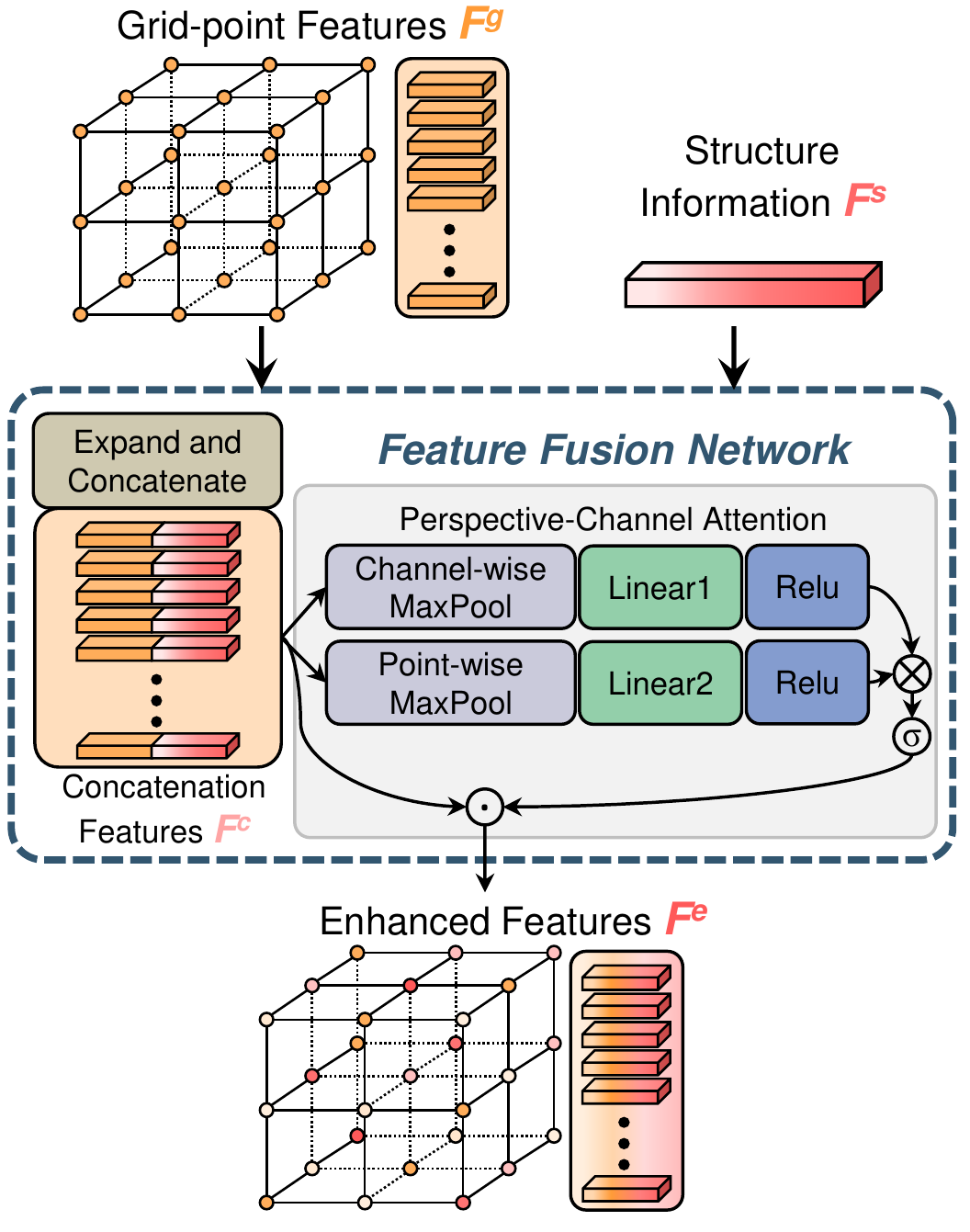}
	\caption{Network architecture of the attention-based feature fusion. The $ \times $, $ \sigma $ and $ \cdot $ denote the matrix multiplication, sigmoid activation and element-wise multiplication respectively.}
	\label{architecture_fusion}
\end{figure}

\section{Architecture of the Feature Fusion Network (Sec 3.2)}
\label{section_atfusion}

\begin{table*}
	\resizebox{\linewidth}{!}{
		\begin{tabular}
			{c ||c c c | c | c c c | c|c c c | c|c c c | c}
			
			\hline
			\multirow{2}{*}{Method} & 
			\multicolumn{4}{c|}{3D Detection (R11)} & 
			\multicolumn{4}{c|}{3D Detection (R40)} &
			\multicolumn{4}{c|}{BEV Detection (R11)} &
			\multicolumn{4}{c}{BEV Detection (R40)} \\
			\cline{2-17}
			& 0-20m & 20-40m & 40m-Inf & Mean & 0-20m & 20-40m & 40m-Inf & Mean & 0-20m & 20-40m & 40m-Inf & Mean & 0-20m & 20-40m & 40m-Inf & Mean\\
			
			\hline
			SECOND \cite{second}& 89.99 & 76.13 & 38.73 & 68.28 & 95.32 & 77.08 & 35.97 & 69.46 & 90.49 & 86.80 & 55.68 & 77.66 & 96.14 & 88.89 & 54.31 & 79.78\\
			
			PointPillars \cite{pointpillars}& 89.60 & 73.28 & 34.14 & 65.67 & 94.46 & 73.01 & 31.07 & 66.18 & 90.24 & 86.08 & 51.16 & 75.83 & 95.63 & 87.66 & 50.66 & 77.98\\
			
			Part-$ A^2 $ \cite{part}& \textbf{90.42} & 76.98 & 39.72 & 69.04 & \textbf{96.30} & 80.03 & 38.01 & 71.45 & \textbf{90.67} & 86.37 & 56.10 & 77.71 & \textbf{96.69} & 88.59 & 54.65 & 79.97\\
			
			PV-RCNN \cite{pvrcnn}& 90.29 & 76.83 & 42.15 & 69.76 & 96.02 & 80.10 & 39.80 & 71.97 & 90.48 & 86.36 & 57.44 & 78.09 & 96.36 & 88.85 & 55.47 & 80.23\\		
			\hline			
			
			SIENet (ours) & 90.33 & \textbf{82.11} & \textbf{46.54} & \textbf{72.99} & 96.23 & \textbf{82.78} & \textbf{44.59} & \textbf{74.53} & 90.60 & \textbf{87.00} & \textbf{59.70} & \textbf{79.10} & 96.67 & \textbf{89.70} & \textbf{60.37} & \textbf{82.25}\\
			
			\rowcolor{MyCyan}\textit{Improvement} & \textit{-0.09} & \textit{+5.13} & \textit{+4.39} & \textit{+3.23} & \textit{-0.07} & \textit{+1.68} & \textit{+4.79} & \textit{+2.56} & \textit{-0.07} & \textit{+0.20} & \textit{+2.26} & \textit{+1.01} & \textit{-0.02} & \textit{+0.81} & \textit{+4.90} & \textit{+2.02}\\
			\hline	
			
	\end{tabular}}
	\caption{The detailed comparison results at different distance ranges on the moderate level car class of KITTI \textit{val} split set. The top performance is shown in bold.}\label{detailed_range_results_table}
\end{table*}

We adopt the perspective-channel attention \cite{3dbonet} to fuse the structure information into the grid point features. As shown in Fig. \ref{architecture_fusion}, the concatenation feature $ \bm{F}^{c} $ is respectively sent to two max pooling layers, \textit{i.e.}, channel-wise and point-wise pooling, to aggregate the features across different dimensions. Two linear layers further encodes the features, and the attention map is generated after the matrix multiplication and sigmoid activate function. Finally, the concatenated feature $ \bm{F}^{c} $ is multiplied with the attention map to generate the enhanced feature $ \bm{F}^{e} \in \mathbb{R}^{216\times (C_1+C_2)} $.

\section{More Quantitative Results on KITTI (Sec 4.2)}
\label{kitti}

In Table. \ref{detailed_range_results_table}, we report a detailed comparison between our SIENet and state-of-the-art methods at different distance ranges. We can see that our method achieves top performance at far object detection, which is consistent with our expectation that the SIENet improves the detection accuracy of far-range objects.

\end{document}